\documentclass[runningheads]{llncs}

\usepackage[table]{xcolor}
\usepackage{eccv}

\usepackage{eccvabbrv}

\usepackage{graphicx}
\usepackage{booktabs}

\usepackage[accsupp]{axessibility}  

\usepackage{hyperref}

\usepackage{orcidlink}

\begin{document}


\title{A Dual-Transformer Architecture with Cross-Attention for Multi-Camera View Recommendation} 

\titlerunning{A Dual-Transformer for Multi-Camera View Recommendation}

\author{Josep Cabacas-Maso\inst{1}\and
Carles Ventura \inst{1} \and Ismael Benito-Altamirano\inst{1,2}}


\institute{eHealth Center, Faculty of Computer Science, Multimedia and Telecommunicactions, Universitat Oberta de Catalunya, 08018 Barcelona, Spain
\email{\{jcabacas,cventuraroy,ibenitoal\}@uoc.edu}\\ \and
Department of Electronic and Biomedical Engineeering, Universitat de Barcelona, 08028 Barcelona, Spain \\
}

\authorrunning{Josep Cabacas-Maso et al.}

\maketitle

\begin{abstract}
  Multi-camera systems are foundational to modern media production, and multi-camera editing is a critical task, this is the proper selection of appropriate camera view at each moment. In this paper, we propose a novel Dual-Transformer architecture with Cross-Attention that heavily outperformed the current SOTA models over the TVMCE dataset (\emph{TV} Shows \emph{M}ulti\emph{c}amera \emph{E}diting dataset). Our model decouples these tasks: (1) a dedicated temporal encoder first processes the sequence of past frames to build a rich memory of the recent history, and (2) the candidate camera views then act as queries to this memory via a cross-attention module, allowing each candidate to independently interrogate the historical context and find the most relevant information for its own evaluation. Our approach achieved 56.60\% Precision@0.5, representing a substantial improvement over the prior best result of 37.16\%. We further conducted an ablation study exploring the use of lightweight backbone architectures, where the SwinV2 backbone yielded the best performance, achieving 69.65\% Precision@0.5. Using this best-performing configuration, we then investigated the feasibility of adapting the model to replicate the editing style of a specific human editor. To this end, we fine-tuned the model using varying proportions of the initial segment of a target video. Our results demonstrate that even with only 20\% of the video used for fine-tuning, the model exhibited measurable improvements in Precision@0.5, indicating strong potential for data-efficient personalization of editing style adapted to each individual TV show or producer. 
  
  \keywords{Multi-View Recommendation \and Cross-Attention \and Dual-Transformer \and Temporal Modeling \and Video Editing \and Deep Learning \and Focal Loss}
\end{abstract}

\section{Introduction}
\label{sec:intro}

Multi-camera systems are foundational to modern media production, playing a significant role in applications ranging from smart city surveillance to professional film production~\cite{Sabha2024}. Within these systems, the task of selecting the appropriate camera view at each moment, often termed multi-camera editing or view recommendation~\cite{Wang2014}, is critical. This process, whether automated or human-driven, directly dictates the final production quality and the audience's narrative experience.

The primary challenge in automating this task lies in modeling the complex, high-dimensional dependencies of a video stream. A successful decision must consider not only the visual content of available camera feeds but also the temporal context of what has been recently shown, and even adhere to established cinematography practices.

\begin{figure}[ht!]
    \centering
    \includegraphics[width=0.9\linewidth]{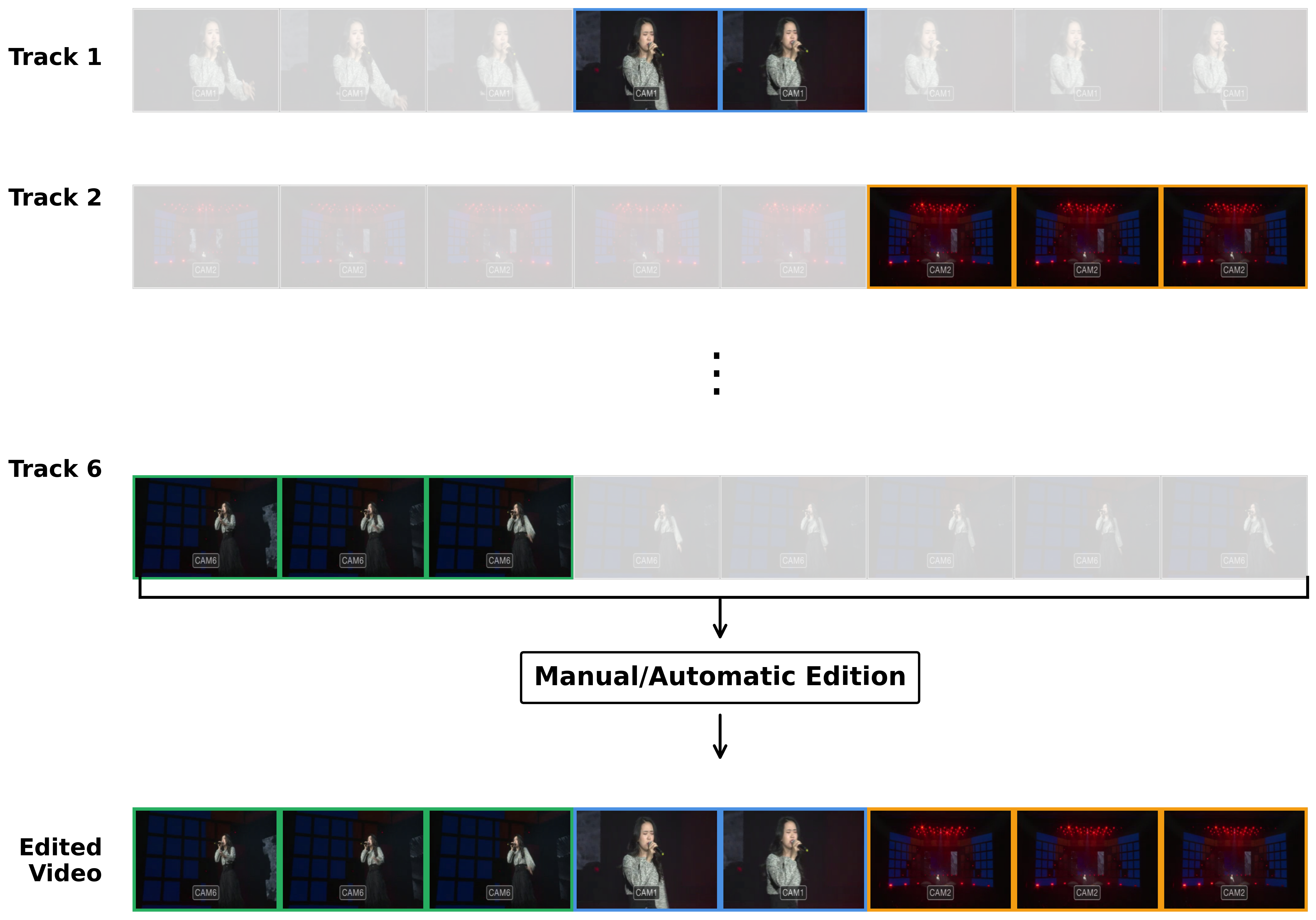}
    \caption{Overview of the multi-camera editing process using an example from the TVMCE dataset~\cite{Rao2022}. A professional editor (or an automated system) selects the most appropriate viewpoint from multiple concurrent camera tracks (shown here as Tracks 1, 2, ..., 6) to construct a final edited sequence that maintains narrative and visual continuity.}
    \label{fig:multi_cam_editing}
\end{figure}

Back in 2014, approaches relied on rule-based systems~\cite{Wang2014}, but the difficulty in capturing these nuanced patterns led to the dominance of data-driven approaches. More recently, in 2022, a pivotal moment in this domain was the introduction of the large-scale TVMCE (TV Shows Multi-Camera Editing) dataset by Rao et al.~\cite{Rao2022}. We consider this is the first benchmark dataset that properly introduces the multi-camera linear editing problem as it is (see~\autoref{fig:multi_cam_editing}). This benchmark, the first of its scale, enabled the advent of deep learning models for the task. Alongside the dataset, Rao et al. proposed the Temporal and Contextual (TC) Transformer, a strong baseline that processed historical shots and candidate views.

This new benchmark spurred further research focused on improving the baseline by incorporating more nuanced features. Currently, the state-of-the-art results for this benchmark were achieved by Lee et al.~\cite{Lee2025} in 2025 by integrating temporal duration and explicit camera identity embeddings into a unified Transformer architecture. This demonstrated that providing the model with features inspired by professional editing practices could significantly boost performance.
However, these existing architectures~\cite{Rao2022,Lee2025} share a common architectural bottleneck. They typically process the historical context and the set of candidate views in a flat or concatenated manner, feeding them as a single sequence into a standard transformer encoder. This \emph{entangles} two distinct cognitive tasks: (1) understanding the temporal dynamics of the \textit{past} and (2) evaluating the set of distinct \textit{future} choices. This is a common challenge in video understanding tasks, which often require reasoning over long temporal horizons~\cite{Wang2016,Shen2022}.
In this paper, we propose a novel \emph{Dual-Transformer architecture with Cross-Attention} that explicitly decouples both tasks into two separate transformer branches. When evaluated on the TVMCE benchmark~\cite{Rao2022}, our model achieves a new SOTA result of 56.60\% Precision@0.5, an improvement of around 20 points over Lee et al.~\cite{Lee2025}, using the same backbone as their own (the SwinV1 transformer~\cite{Liu2021_Swin}), and up to 69.95\% Precision@0.5 using the SwinV2 transformer~\cite{Liu_2022_CVPR}. Finally, we demonstrated that our model can \emph{imitate} a targeted human editor by fine-tuning over a target video edition using only 20\% of the target video editions to achieve an 83.89\% Precision@0.5 for the test subset of that video.

\section{Related Work}

Our work builds upon three main pillars of research: automated video editing, deep learning for multi-camera view recommendation, and the application of transformer architectures to video understanding.

\subsection{Automated Video Editing}

Automating video editing is a long-standing goal in computer vision and multimedia fields. Research in this area is diverse, tackling sub-problems such as: video summarization~\cite{Wang2020_trailer,Zhang2016_summary}, highlight detection~\cite{Evangelidis2014_highlight}, or automatic recommendation of transition effects~\cite{Shen2022}. Much of this work has focused on domain-specific editing rules. For example, several approaches have been developed for automatically editing live sports broadcasts~\cite{Hsieh2014_sport} which often follow predictable patterns. Other works have focused on editing theater recordings~\cite{Stoll2023_Access}. In the domain of virtual camera systems, Achary et al.~\cite{Achary_2024_WACV} proposed \textit{Real Time GAZED}, which selects shots from monocular wide-angle recordings using gaze data and a look-ahead energy-minimization strategy

The main problem that automated video editing pose in order to train proper models is the lack of actual edited videos with the corresponding raw footage. Authors have tackled a way to circumvent this by pseudo-labeling a dataset by decomposing edited videos from elsewhere. For example, Jimenez et al.~\cite{jimenez2022automated} created a dataset of concerts and other theatrical performances by decomposing edited videos from YouTube into an approximation of the raw footage, taking frames from other parts of the same video as the ``raw'' footage of other confuser cameras. Despite their efforts their dataset was unable to properly capture the multi-camera editing problem as it is, because their choice as confuser frames was flawly based on selecting frames from the same video or other videos from the same dataset.

Later, Lee et al.~\cite{lee2024pseudo} standardized a way of pseudo-labeling multi-camera dataset by decomposing videos from elsewhere, like~\cite{jimenez2022automated}, but using a more robust method to select the confuser frames, by clustering the decomposed video views into pseudo camera-views. More recently, this same methodology was refined by Gonzálbez-Biosca~\cite{Gonzalbez2024_concert} in a more relevant work for domain-specific classical musicals, where they make a key architectural contribution to decompose the problem into two distinct sub-tasks:``when to cut'' (temporal segmentation) and ``how to cut'' (camera selection).

\subsection{Multi-Camera Editing and View Recommendation}

As before-explained, the specific task of multi-camera editing, or view recommendation, involves selecting the best viewpoint from a set of synchronized cameras at any given moment, and the lack of actual data is crucial to properly evaluate the problem. Early data-driven approaches were often limited by small-scale datasets~\cite{Wang2014} or relied on rule-based systems that were difficult to scale~\cite{Wang2016}.

The field was significantly advanced by the introduction in the past years of the large-scale TVMCE dataset by Rao et al.~\cite{Rao2022}. To address the scarcity of high-quality data, they collected professionally produced footage from diverse scenarios, including concerts, sports, and gala shows. The raw footage was recorded by trained videographers and subsequently edited by professional directors. This process ensured that the dataset captured realistic, high-quality editing decisions. A key challenge they overcame was the synchronization of multiple camera tracks, which is essential for the task. The final dataset, TVMCE, spans 88 hours of raw footage, resulting in 14 hours of edited video, providing a robust foundation for training and evaluating deep learning models.

This dataset enabled the development of robust deep learning models, although it is worth noting that while the dataset is extensive, it is not multimodal. This limits the applicability of models that rely on other modalities, such as audio, to determine when to change the camera~\cite{Gonzalbez2024_concert}. Their baseline, the Temporal and Contextual (TC) Transformer, demonstrated the feasibility of using a Transformer to learn editing decisions by jointly considering historical shots and candidate views.

Subsequent work, to outperform their original baseline, has aimed to incorporat more explicit cinematographic knowledge to the models. For example, Lee et al.~\cite{Lee2025} achieved the current best state-of-the-art result by integrating temporal duration and camera identity as learned embeddings. They showed that providing the model with features inspired by professional editing heuristics improves performance. Similarly, other works have explored using features like shot type classification~\cite{Lei2020} to guide video analysis.

\subsection{Transformers in Video Understanding}

Out of the scope of multi-camera editing, transformer architectures have become the \emph{de facto} standard for sequence modeling~\cite{Vaswani2017}, and their application has extended deep understanding of video edition for expert recommendation systems. Models such as the I3D~\cite{Carreira2017} and Temporal Segment Networks (TSN)~\cite{Wang2016} have established strong baselines for video action recognition. More recently, pure-transformer architectures like the Video Vision Transformer (ViViT)~\cite{Arnab2021} and SwinV1~\cite{Liu2021_Swin} and SwinV2~\cite{Liu_2022_CVPR} transformers have pushed the boundaries of video classification.


\section{Proposed Method}



\begin{figure}[htbp]
  \centering
  \includegraphics[width=0.95\textwidth]{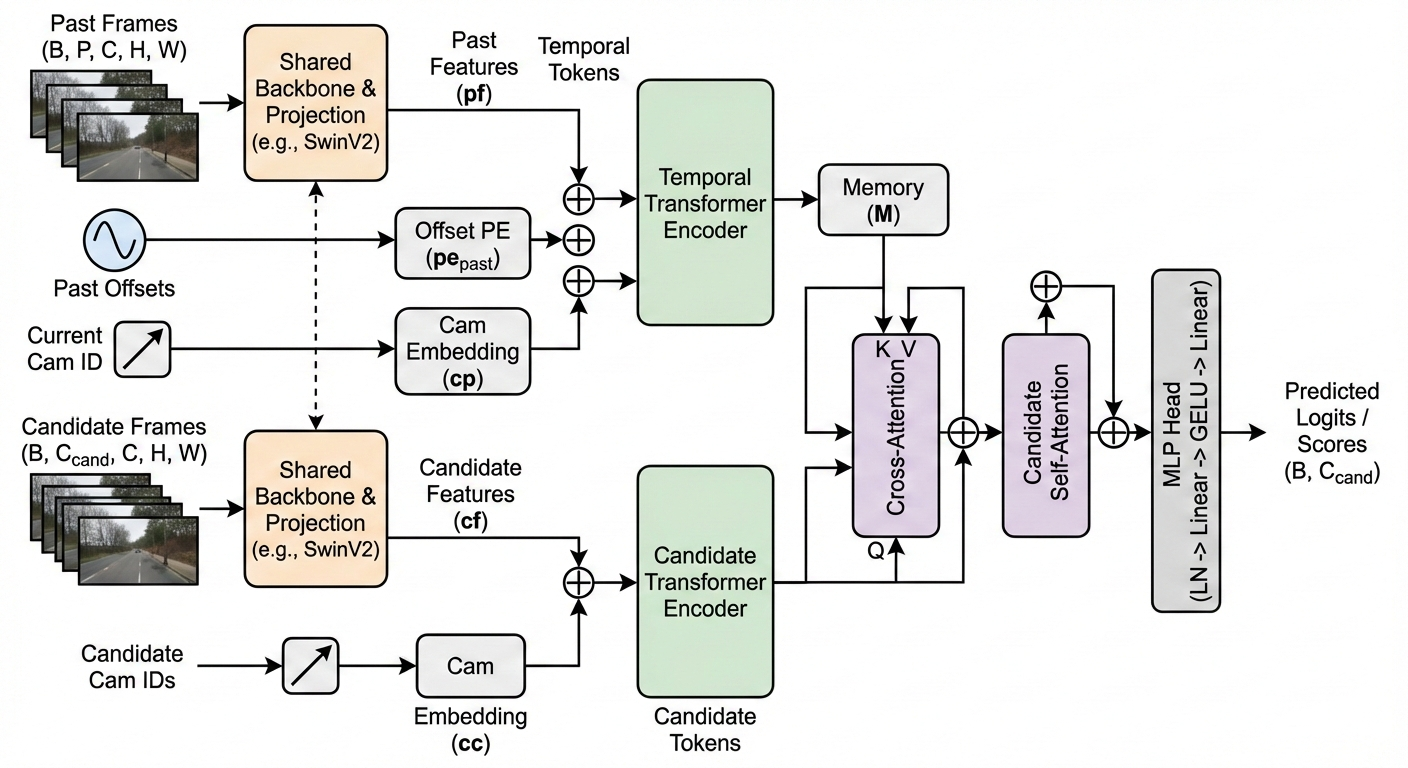}
  \caption{Overview of the proposed Dual-Transformer architecture. The model decouples temporal context encoding from candidate view evaluation. Past frames are processed through a shared backbone (e.g.~\texttt{SwinV2}) and integrated with temporal and camera embeddings via a \texttt{Temporal Transformer Encoder}. A \texttt{Cross-Attention} mechanism then aligns these historical memory features with candidate frame features, followed by a \texttt{Candidate Self-Attention} layer and a \texttt{Multi-Layer Perceptron (MLP) Head} to produce final recommendation scores.}
  \label{fig:architecture}
\end{figure}

As outlined in~\autoref{sec:intro}, previous methods process historical and
candidate-view information in an entangled manner. We propose a
\textbf{Dual-Transformer architecture with Cross-Attention} that explicitly
decouples the task into two stages (see~\autoref{fig:architecture}). First, a dedicated \textbf{Temporal Encoder} processes the sequence of $P$ past frames to build a rich, contextual memory of the recent history, drawing inspiration from advances in video action recognition~\cite{Carreira2017,Arnab2021}. Second, the candidate camera views, enriched with their own features, act as \textit{queries} to this memory. This is achieved via a \textbf{Cross-Attention} module~\cite{Vaswani2017}, allowing each candidate to independently query the historical context and find the most relevant information for its own evaluation.

\subsection{Feature and Positional Encoding}
Before being processed by the Transformer blocks, the raw inputs are converted into a sequence of $D$-dimensional feature tokens ($D=768$).

\textbf{Visual Feature Extraction.} We employ a pre-trained visual backbone, such as the \textbf{Swin Transformer V2}~\cite{Liu_2022_CVPR}, to extract high-level semantic features. For a sequence of $N$ input images, the backbone outputs visual feature vectors $F \in \mathbb{R}^{N \times D_{backbone}}$. We apply a linear projection layer to map these features to our model's internal dimension $D$.

\textbf{Camera Identity Embedding.} Following~\cite{Lee2025}, we use a learnable embedding layer to represent camera identities. For a camera index $i$, the model retrieves an embedding $E_{cam, i} \in \mathbb{R}^{D}$.

\textbf{Temporal Offset Encoding.} To represent the temporal distance of past frames, we use sinusoidal positional encodings $PE_{offset}$ based on integer frame offsets $o$, as in~\cite{Lee2025}. Offsets are clamped at a maximum value of 500:
\begin{align}
    PE_{offset}(o, 2j) &= \sin(o / 10000^{2j/D}) \\
    PE_{offset}(o, 2j+1) &= \cos(o / 10000^{2j/D})
\end{align}

\subsection{Dual-Transformer Architecture}
Our architecture consists of two symmetric Transformer-based pathways that converge in a cross-attention module.

\subsubsection{Temporal Encoder (Memory)}
The Temporal Encoder processes $P=16$ past tokens $T_{past} \in \mathbb{R}^{P \times D}$, formed by the element-wise sum of visual features, temporal encodings, and the current camera identity:
\begin{equation}
    T_{past} = F_{past} + PE_{offset} + E_{cam, curr}
\end{equation}
These tokens pass through a Temporal Transformer Encoder to generate the historical memory $M \in \mathbb{R}^{P \times D}$, which serves as the Key ($K$) and Value ($V$) for the subsequent stage.

\subsubsection{Candidate Encoding and Cross-Attention}
Crucially, we do not concatenate candidates with the history. Instead, we create $C$ candidate tokens $T_{cand} \in \mathbb{R}^{C \times D}$ by combining visual features with their respective camera identity embeddings:
\begin{equation}
    T_{cand} = F_{cand} + E_{cam, cand}
\end{equation}
These tokens are first processed by a \textbf{Candidate Transformer Encoder} to capture context among the available views. The resulting tokens form the Query ($Q$). We then apply Multi-Head Cross-Attention :
\begin{equation}
    T'_{cand} = T_{cand} + \text{CrossAttn}(Q=T_{cand}, K=M, V=M)
\end{equation}
Finally, a shallow self-attention layer allows the candidates to interact after fetching historical context, producing the final refined tokens $T''_{cand}$.

\subsection{Prediction Head and Loss Function}
The final tokens are passed through an MLP head with Layer Normalization and GELU activation to produce logits. To handle the class imbalance, we employ a \textbf{Focal Loss}~\cite{Lin2017_FocalLoss}. Given the logit $z$ and target $y \in \{0, 1\}$, our implementation uses:
\begin{equation}
    FL(z, y) = \alpha_t (1 - p_t)^\gamma \cdot BCE(z, y)
\end{equation}
where $p_t$ is the sigmoid probability corresponding to the correct class, with hyper-parameters $\alpha=0.25$ and $\gamma=2.0$ as defined in our training logic.

\section{Experiments and Analysis}
We conduct an extensive empirical evaluation to validate our Dual-Transformer architecture. Our experiments are designed to address three primary research questions: (i) Does the decoupling of temporal memory from candidate queries improve view recommendation? (ii) Which visual backbones provide the most discriminative features for this task? and (iii) What is the minimum volume of data required to predict an individual editor's camera-selection behavior?

\subsection{Experimental Setup}

\textbf{Dataset.} We train and evaluate our model on the large-scale TVMCE dataset~\cite{Rao2022}. This dataset contains 88-hour raw videos that contribute to the 14-hour edited videos, spanning concerts, sports, gala shows, and contests. The task involves predicting the optimal camera view at each editing boundary. Following the benchmark protocol, the model is provided with a history of $P=16$ past shots and a set of $C=6$ candidate views.

\textbf{Implementation Details.} Our framework is implemented\footnote{%
Our code is released at
\url{https://gitlab.com/UOC/aiwell/automated-video-editing-research/eccv2026-dualtransformer}.} in PyTorch. For the visual backbone, we utilize a Swin Transformer V2-Tiny~\cite{Liu_2022_CVPR} pre-trained on ImageNet-1K. To maintain computational efficiency and focus on architectural reasoning, the backbone weights are frozen, and features are projected to a latent dimension of $D=768$ via a linear layer. 

The core architecture consists of a 1-layer Transformer for the Temporal Encoder and a symmetric 1-layer Transformer for Candidate Encoding. The Cross-Attention module utilizes $H=8$ attention heads, while the post-attention refinement is handled by a 4-head self-attention layer. The prediction head is a two-layer MLP (192-node hidden layer) with GELU activations. Training is conducted for 30 epochs using the AdamW optimizer with a learning rate of $1 \times 10^{-5}$. To address class imbalance, we apply Focal Loss~\cite{Lin2017_FocalLoss} with $\alpha=0.25$ and $\gamma=2.0$.

\textbf{Evaluation Metric.} Consistent with prior literature~\cite{Rao2022,Lee2025}, our primary evaluation metric is \textbf{Precision@0.5 (P@0.5)}. This metric evaluates the model's ability to identify the correct shot while successfully suppressing the five distractor candidates. A prediction is classified as a True Positive (TP) only if its sigmoid probability exceeds the 0.5 threshold and it is the correct shot to be selected. 

In our evaluation protocol, precision is computed independently for each sample\footnote{A sample refers to the output of the model for a given sequence of P past frames and the selection among C view candidates.} and then averaged across all samples. This means that every prediction instance contributes equally to the final metric, regardless of how many positive labels are activated for that sample. Notice that, for each sample, we may have more than one positive sample (sigmoid probability greater than a given threshold) but, at most, only one becomes a TP. Formally, if $N$ denotes the total number of samples, the reported precision corresponds to:
\begin{equation}
\text{Precision}_{macro} = \frac{1}{N} \sum_{i=1}^{N} \frac{TP_i}{TP_i + FP_i}
\end{equation}
This macro-averaged formulation ensures that each prediction has the same weight in the final score, preventing samples with a large number of predicted labels from disproportionately influencing the metric. 

\subsection{Main Results: Establishing a New State-of-the-Art}
We benchmark our model against several key baselines on the TVMCE dataset. These include the original \textbf{TC-Transformer}~\cite{Rao2022}, which utilizes a unified encoder, and \textbf{Lee et al.}~\cite{Lee2025}, the previous state-of-the-art that improved performance by adding temporal and camera embeddings to a single-transformer backbone. We also include the multimodal concert editing framework of \textbf{Gonzálbez-Biosca et al.}~\cite{Gonzalbez2024_concert}.

\begin{table}[ht]
\centering
\caption{Main results on the TVMCE benchmark. Precision@0.5 (\%) is used as the primary metric. Our Dual-Transformer architecture with SwinV2 backbone establishes a significant new performance upper bound. To isolate the architectural contribution from the backbone, we additionally re-implement the prior SOTA~\cite{Lee2025} under both backbones.}
\label{tab:main_results}
\begin{tabular}{@{}llc@{}}
\toprule
\textbf{Category} & \textbf{Model} & \textbf{Precision@0.5 (\%)} \\
\midrule
Baselines & Random Chance & 16.67 \\
& Gonzálbez-Biosca et al.~\cite{Gonzalbez2024_concert} & 17.65 \\
& TC-Transformer~\cite{Rao2022} & 22.50 \\
& Lee et al.~\cite{Lee2025} (Previous SOTA) & 37.16 \\
\midrule
Re-impl.\ SOTA$^{\dagger}$ & Single Transformer~\cite{Lee2025} (BCE + SwinV1) & 47.97 \\
& Single Transformer~\cite{Lee2025} (BCE + SwinV2) & 54.06 \\
\midrule
\textbf{Ours} & \textbf{Dual-Transformer (BCE + SwinV1)} & 52.60 \\
& \textbf{Dual-Transformer (Focal + SwinV1)} & 56.60 \\
& \textbf{Dual-Transformer (Focal + SwinV2)} & \textbf{69.65} \\
\bottomrule
\end{tabular}
\\[4pt]
{\footnotesize $^{\dagger}$Since the prior SOTA~\cite{Lee2025} provides no public code or model weights, these figures represent our most faithful re-implementation based on the details reported by the authors.}
\end{table}

The empirical evidence in Table~\ref{tab:main_results} demonstrates a paradigm shift in performance. By simply introducing the decoupled architecture (\textit{Ours-BCE Loss}), we observe a Precision@0.5 of 52.60\%, which already exceeds the previous SOTA by a wide margin. This validates our hypothesis that a unified encoder suffers from feature entanglement, where historical context and future candidates are mixed prematurely. 

\begin{figure}[h!]
  \centering
  \includegraphics[width=0.9\textwidth]{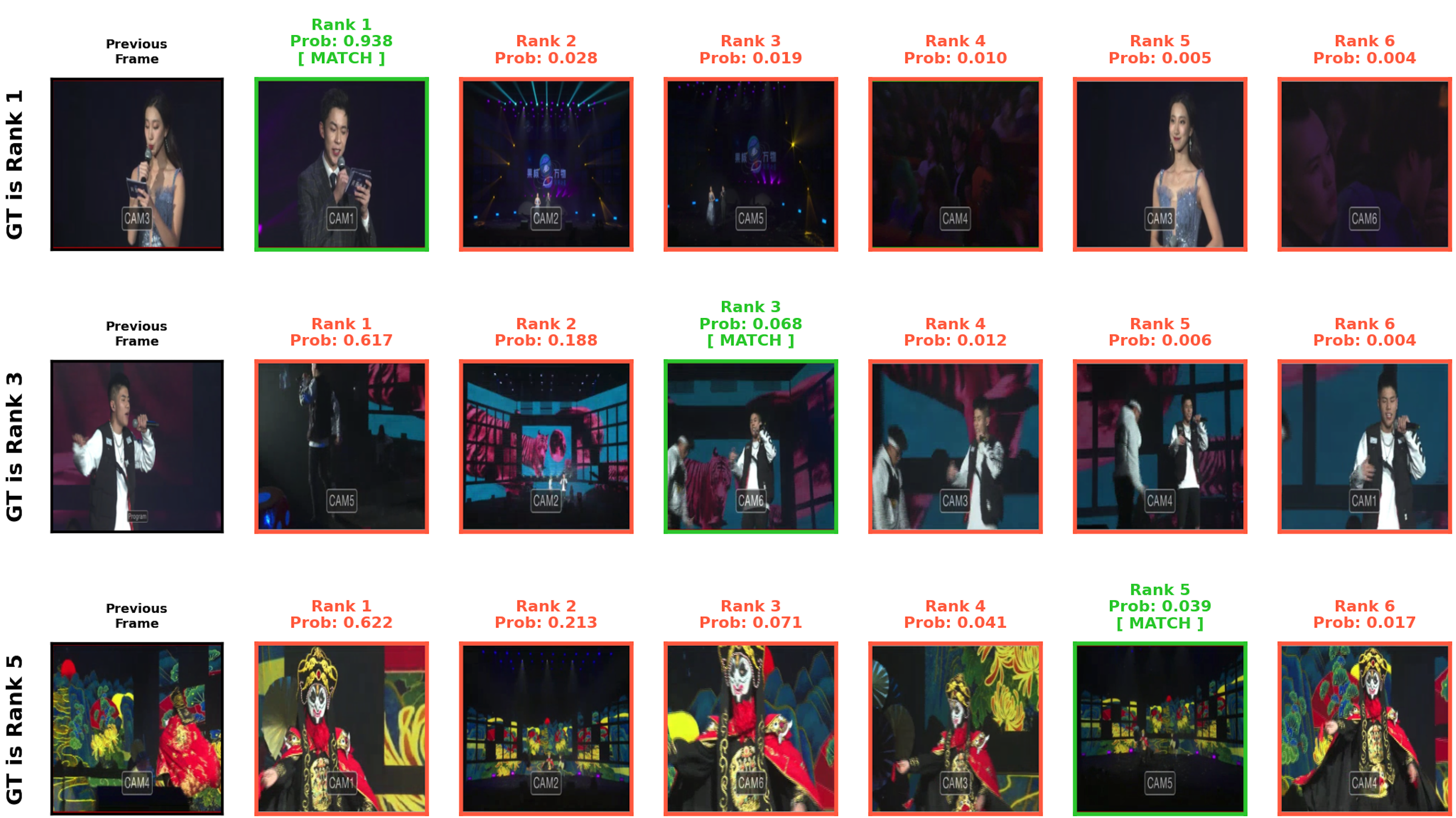}
  \caption{Sample predictions from the best-performing model (Dual-Transformer with SwinV2-Tiny backbone and Focal Loss). For each row, the first element is the last frame from the past  frames and the remaining elements are the ranking of the C candidate views given by the model with the associated confidence score. The candidate framed in green is the ground truth. Top: Best candidate matches with ground truth (P@0.5=1). Middle: Ground truth frame is placed at 3rd position (P@0.5=0). Bottom: Ground truth fame is placed at 5th position (P@0.5=0).}
  \label{fig:predictions}
\end{figure}

The integration of \textbf{Focal Loss}~\cite{Lin2017_FocalLoss} further optimizes the model for the sparse signal inherent in editing (where one correct shot exists among many negatives) by achieving a Precision@0.5 of 56.60\%. Finally, our flagship configuration using the \textbf{SwinV2-Tiny}~\cite{Liu_2022_CVPR} backbone and including the Focal Loss reaches a Precision@0.5 of 69.65\%. This performance level suggests that the model has moved beyond simple statistical correlation and is beginning to learn the underlying ``cinematographic language'' of professional editors. \autoref{fig:predictions} shows some qualitative results.

\subsection{Ablation Study: The Impact of Visual Backbones}
To isolate the contribution of the visual backbone from the architectural design, we performed a systematic ablation using our Dual-Transformer framework as the constant. We tested different backbones across convolutional, vanilla transformer, and hierarchical transformer families.

\begin{table}[h]
\centering
\caption{Backbone sensitivity analysis. All models utilize our Dual-Transformer with Focal Loss. Results highlight the superiority of hierarchical window-based attention for high-resolution editing cues.}
\label{tab:backbones}
\begin{tabular}{lc|lc}
\toprule
\textbf{Backbone} & \textbf{Precision@0.5 (\%)} & \textbf{Backbone} & \textbf{Precision@0.5 (\%)} \\
\midrule
\textbf{SwinV2-Tiny~\cite{Liu_2022_CVPR}} & \textbf{69.65} & ConvNeXt-Tiny & 40.82 \\
MaxViT-Tiny & 58.63 & EfficientNet-B0 & 40.66 \\
SwinV1-Tiny~\cite{Liu2021_Swin} & 56.60 & CoAtNet-0 & 34.42 \\
ConvNeXtV2-Tiny & 56.23 & EVA02-Tiny & 33.95 \\
ResNet-50 & 46.74 & ViT-Base~\cite{Vaswani2017} & 25.85 \\
\bottomrule
\end{tabular}
\end{table}

The data in Table~\ref{tab:backbones} reveals several critical insights:
\begin{enumerate}
    \item \textbf{Hierarchy Matters:} Hierarchical models (Swin, MaxViT) consistently outperform non-hierarchical Vision Transformers (ViT-Base). The poor performance of ViT-Base (25.85\%) indicates that standard global attention may lose the fine-grained spatial details necessary for distinguishing between similar camera angles.
    \item \textbf{Window-based Attention:} SwinV2's shift to log-spaced positional bias and shifted windows allows for better resolution of candidate views, achieving 69.65\%.
    \item \textbf{CNN Resilience:} While modern CNNs like ConvNeXtV2 perform admirably (56.23\%), they fail to match the cross-view reasoning capabilities of the Swin family.
\end{enumerate}

\subsection{Extending the evaluation protocol}

Following the analytical framework in~\cite{cabacas2024enhancing}, we conducted a sensitivity analysis on the decision threshold $\tau$. While $\tau=0.5$ is the established benchmark for comparability, the nature of view recommendation problem warrants a broader evaluation of the precision-recall trade-off. We introduce Recall and $F_1$-score to provide a holistic measure of the model's reliability.

\begin{table}[h!]
\centering
\caption{Threshold selection study on the validation set vs. final performance on the test set (Macro-Averaged). The optimal threshold $\tau=0.3$ (identified via validation $F_1$-score) is used for the final test evaluation.}
\label{tab:threshold_study_macro}
\begin{tabular}{@{}cccc@{}}
\toprule
\textbf{Threshold ($\tau$)} & \textbf{Precision (\%)} & \textbf{Recall (\%)} & \textbf{$F_1$-Score (\%)} \\ \midrule
\multicolumn{4}{c}{\textit{Validation Set Sweep}} \\ \midrule
0.1 & 67.81 & 95.21 & 75.74 \\
0.2 & 76.87 & 90.59 & 81.21 \\
\rowcolor[HTML]{EFEFEF} \textbf{0.3} & \textbf{79.41} & \textbf{85.48} & \textbf{81.41} \\
0.4 & 78.51 & 80.28 & 79.10 \\
0.5 & 74.46 & 74.67 & 74.53 \\
0.6 & 68.52 & 68.56 & 68.54 \\
0.7 & 58.99 & 58.99 & 58.99 \\
0.8 & 47.77 & 47.77 & 47.77 \\
0.9 & 30.86 & 30.86 & 30.86 \\ \midrule
\multicolumn{4}{c}{\textit{Final Performance (Test Set)}} \\ \midrule
\textbf{0.3 (Optimal)} & \textbf{74.66} & \textbf{80.35} & \textbf{76.52} \\ \bottomrule
\end{tabular}
\end{table}

\noindent
As illustrated in Table \ref{tab:threshold_study_macro}, we conducted a grid search on the validation set to determine the decision threshold $\tau$ that best balances recommendation accuracy and coverage. The model achieves its peak validation $F_1$-score of $81.41\%$ at $\tau=0.3$. We therefore fix this threshold for our final evaluation on the unseen test set, where the model maintains strong performance with a Precision of $74.66\%$, a Recall of $80.35\%$, and an $F_1$-score of $76.52\%$.

The observation that the optimal threshold ($\tau=0.3$) is lower than the standard $0.5$ suggests that the Dual-Transformer architecture produces well-separated but ``under-confident'' probability distributions. As shown in the validation sweep, increasing the threshold to $0.5$ leads to a sharp decline in the validation $F_1$-score (from $81.41\%$ to $74.53\%$) because the model assigns moderate confidence to a large number of correct views. This behavior is a known characteristic of models trained with Focal Loss, which prioritizes the relative ranking of hard examples over absolute probability calibration. By tuning $\tau$ on the validation set, we successfully recover these accurate predictions that would otherwise be discarded by a default threshold.

\bigskip

\noindent
\textbf{Micro-Averaged Evaluation.} In addition to the instance-balanced (macro-averaged) evaluation described above, we also performed a global (micro-averaged) computation of the metrics. In this formulation, True Positives, False Positives, and False Negatives are accumulated across the entire dataset before computing the evaluation metrics. Formally:

\begin{equation}
\text{Precision}_{micro} = \frac{\sum_i TP_i}{\sum_i TP_i + \sum_i FP_i}
\end{equation}

Unlike the macro-averaged formulation where each sample contributes equally to the final score the micro-averaged computation assigns equal weight to each individual prediction decision. Consequently, samples producing a larger number of positive predictions have a stronger influence on the final metric.

\begin{table}[h!]
\centering
\caption{Threshold selection study using micro-averaged (global) Precision, Recall, and $F_1$-score. The optimal threshold $\tau=0.4$ (identified via validation $F_1$-score) is used for the final test evaluation.}
\label{tab:threshold_micro}
\begin{tabular}{@{}cccc@{}}
\toprule
\textbf{Threshold ($\tau$)} & \textbf{Precision (\%)} & \textbf{Recall (\%)} & \textbf{$F_1$-Score (\%)} \\ \midrule
\multicolumn{4}{c}{\textit{Validation Set Sweep}} \\ \midrule
0.1 & 54.90 & 95.21 & 69.65 \\
0.2 & 68.33 & 90.59 & 77.90 \\
0.3 & 76.01 & 85.48 & 80.47 \\
\rowcolor[HTML]{EFEFEF} \textbf{0.4} & \textbf{81.02} & \textbf{80.28} & \textbf{80.65} \\
0.5 & 85.62 & 74.67 & 79.77 \\
0.6 & 89.94 & 68.56 & 77.81 \\
0.7 & 92.74 & 58.99 & 72.11 \\
0.8 & 95.23 & 47.77 & 63.63 \\
0.9 & 97.14 & 30.86 & 46.84 \\ \midrule
\multicolumn{4}{c}{\textit{Final Performance (Test Set)}} \\ \midrule
\textbf{0.4 (Optimal)} & \textbf{77.52} & \textbf{74.75} & \textbf{76.11} \\ \bottomrule
\end{tabular}
\end{table}

\noindent
The shift in optimal threshold from $\tau=0.3$ (macro) to $\tau=0.4$ (micro) is a consequence of the different weighting strategies. Micro-averaging penalizes false positives globally, favoring slightly higher thresholds to improve overall prediction-level precision. Macro-averaging treats each sample equally, recovering more correct labels per instance even at lower thresholds. Despite this difference, both approaches yield comparable test $F_1$-scores, confirming the robustness of the model and the stability of the threshold tuning process.

\noindent
\textbf{Evaluation using Recall@1.} Regarding the evaluation metrics, the view recommendation problem can be also considered as an information retrieval problem, where there are other metrics such as Recall@K, which computes the ratio of relevant items among the top K candidates in a rank list.

In the context of the view recommendation problem, there is only one possible relevant item per sample. Therefore, we decided also to consider the strictest evaluation metric, i.e. Recall@1, which corresponds to the percentage of samples where the most confident candidate matches with the ground truth. Notice that there is no need to consider any threshold to compute this metric. Our proposed model achieves a Recall@1 of \textbf{76.31\%}. This confirms that the model reliably selects the most confident camera for each sample, providing a clear baseline for top-1 selection performance.




\subsection{Data Efficiency and Cross-Video Adaptation}
A primary challenge in automated video editing is the ability of a model to adapt to the specific stylistic nuances of a professional editor with minimal supervision. To evaluate the adaptability of our Dual-Transformer architecture, we use two high-resolution sequences from the test set: \textit{video\_0000} and \textit{video\_0001}.

\textbf{Experimental Protocol.} We treat each video as an independent domain. For a given video, we split the data into local train, validation, and test sets. We begin with a \textbf{zero-shot} evaluation, where our best-performing model (trained on the general TVMCE training set) is applied directly to the test split of the target video. Subsequently, we fine-tune the model using incremental percentages ($10\%$, $20\%$, and $30\%$) of the video's local training data. To test for cross-video generalization, we evaluate the model trained on \textit{video\_0000} on the test set of \textit{video\_0001}, and vice versa.

\begin{table}[ht]
\centering
\caption{Cross-video fine-tuning results. We report Precision@0.5 (\%) across different training data percentages.} 
\label{tab:adaptation_results}
\begin{tabular}{@{}lccccc@{}}
\toprule
\textbf{Source Video} & \textbf{Data \%} & \textbf{Target Video} & \textbf{Precision@0.5 (\%)} \\
\midrule
General TVMCE & 0\%  & video\_0000 & 74.44 \\
General TVMCE & 0\%  & video\_0001 & 81.76 \\
\midrule
video\_0000 & 10\% & video\_0000 & 78.76 \\
video\_0000 & 10\% & video\_0001 & 78.82 \\
video\_0000 & 20\% & video\_0000 & \textbf{83.65} \\
video\_0000 & 20\% & video\_0001 & 76.47 \\
video\_0000 & 30\% & video\_0000 & 90.41 \\
video\_0000 & 30\% & video\_0001 & 88.82 \\
\midrule
video\_0001 & 10\% & video\_0001 & 77.06 \\
video\_0001 & 10\% & video\_0000 & 80.64 \\
video\_0001 & 20\% & video\_0001 & \textbf{84.12} \\
video\_0001 & 20\% & video\_0000 & 83.65 \\
video\_0001 & 30\% & video\_0001 & 86.18 \\
video\_0001 & 30\% & video\_0000 & 85.71 \\
\bottomrule
\end{tabular}
\end{table}

\textbf{Analysis of Human-Editor Reproducibility.} 
The results in Table~\ref{tab:adaptation_results} reveal three significant findings:
\begin{enumerate}
    \item \textbf{High Zero-Shot Baseline:} Our general model achieves over 74\% precision on unseen videos without fine-tuning, indicating that the Dual-Transformer architecture captures universal cinematographic principles (e.g., standard shot-reverse-shot patterns) from the broader dataset.
    
    \item \textbf{The 20\% Inflection Point:} Using only \textbf{20\% of the training data} from a specific video, the model reaches approximately 84\% precision. This pivotal result demonstrates that the model can learn the specific rhythm and view preference of a human editor from a fraction of a single production's data, making the system viable for real-world environments where data labeling is expensive.
    
    \item \textbf{Cross-Video Robustness:} A model fine-tuned on \textit{video\_0001} and evaluated on \textit{video\_0000} remains remarkably accurate (83.65\% at the 20\% mark). This stability suggests the fine-tuned features are not overfitting to the visual artifacts of one video, but refining the model's understanding of camera relationships consistent across similar production types.
\end{enumerate}
These findings confirm that decoupling temporal history and candidate selection provides the structural prior needed for rapid, high-fidelity adaptation to human editing styles.

\section{Conclusion}
In this work, we addressed a fundamental limitation in automated multi-camera view recommendation: the structural entanglement of temporal history and candidate view evaluation within unified Transformer architectures. To resolve this, we introduced a novel \textbf{Dual-Transformer architecture} that explicitly decouples the reasoning process into a dedicated temporal memory and a cross-attention module.
Our empirical evaluation on the TVMCE benchmark demonstrates that this architectural shift, combined with a \textbf{Swin Transformer V2} backbone and Focal Loss optimization, establishes a new state-of-the-art, achieving a Precision@0.5 of \textbf{69.65\%}. Our ablation studies across diverse visual backbones further confirm that hierarchical window-based attention provides the most discriminative features for cinematographic decision-making.
Our investigation into diverse evaluation metrics revealed the model's versatility. By optimizing decision thresholds ($\tau=0.3$ macro; $\tau=0.4$ micro), we recovered accurate predictions that would otherwise be discarded, yielding a test F1-score of \textbf{76.52\%}, and a Recall@1 of \textbf{76.31\%} when considering only the most confident camera. This confirms the model balances the precision-recall trade-off well and is reliable for fully automated view recommendation.
Finally, we demonstrated the data efficiency of our approach: achieving over 83\% Precision@0.5 with only \textbf{20\% of the target video's data}, our model proves capable of rapidly reproducing the subtle editing patterns of human professionals. 
Future work will explore integrating multimodal cues such as audio dynamics and actor movement to further refine the ``when to cut'' logic, moving toward a fully autonomous, professional-grade directorial system.
\section*{Acknowledgements}

This work has been partially supported by Grant PID2022-138721NB-I00, funded by MCIN/AEI/10.13039/501100011033 and FEDER,EU. I. Benito-Altamirano acknowledges the support from 2025-DI-00035 funded by the DI Plan from AGAUR.

%
%
\bibliographystyle{splncs04}
\bibliography{main}
\end{document}